\PassOptionsToPackage{dvipsnames,table}{xcolor}
\newif\ifarxiv
\arxivtrue 

\documentclass[letterpaper,10pt,conference]{IEEEtran}
\IEEEoverridecommandlockouts
\usepackage{xcolor}
\usepackage{amsmath,amssymb,graphicx,url}
\usepackage{tikz}
\usetikzlibrary{arrows.meta,positioning}
\usepackage[numbers,sort&compress]{natbib}

\usepackage{upgreek}
\usepackage{multirow}
\newcommand{\review}[1]{\textcolor{red}{#1}}
\newcommand{\problem}[1]{\par\noindent\textbf{Problem.} #1}
\newcommand{\runin}[1]{\par\indent\textit{#1.}\enspace}
\newtheorem{assumption}{Assumption}

\newcommand{\methodLongName}{Split Label-Localized Conformal Prediction}
\newcommand{\methodShortName}{SLLCP}
\newcommand{\methodFullName}{\methodLongName{} (\methodShortName{})}

\newcommand{\methodRawBinaryName}{Raw VLM (binary)}
\newcommand{\methodRawBandName}{Raw VLM ($0.1/0.9$)}
\newcommand{\methodNoLocalityName}{Ablation w/o locality}
\newcommand{\methodNoProbeName}{Ablation w/o logistic probe}
\newcommand{\methodNoProjectionName}{Ablation w/o projection}
\newcommand{\methodNoLabelCPName}{Ablation w/o label-conditional CP}
\newcommand{\methodOursName}{\methodShortName{} (ours)}

\newcommand{\state}{x}
\newcommand{\horizon}{H}

\newcommand{\monitor}{m}
\newcommand{\predProb}{\hat{p}}
\newcommand{\safeLabel}{\textsc{safe}}
\newcommand{\unsafeLabel}{\textsc{unsafe}}

\newcommand{\vlmEmbedding}{\mathbf{e}_{\mathrm{VLM}}}
\newcommand{\projectionMap}{g}
\newcommand{\projectedEmbedding}{\mathbf{z}}

\newcommand{\CPinput}{X}
\newcommand{\CPoutput}{Y}
\newcommand{\CPpred}{\widehat{Y}}
\newcommand{\CPcandidate}{y}
\newcommand{\CPinputSpace}{\mathcal{X}}
\newcommand{\CPoutputSpace}{\mathcal{Y}}

\newcommand{\calibset}{D_{\mathrm{cal}}}
\newcommand{\localCalSet}{D_{\mathrm{cal}}^{\mathrm{local}}}
\newcommand{\cpCalSet}{D_{\mathrm{cal}}^{\mathrm{CP}}}

\newcommand{\numcal}{n}
\newcommand{\CPinputTest}{\CPinput_{\numcal+1}}
\newcommand{\CPoutputTest}{\CPoutput_{\numcal+1}}

\newcommand{\scoreFunc}{r}
\newcommand{\scoreVal}{R}

\newcommand{\localIndexSet}{I_{\mathrm{local}}}
\newcommand{\cpIndexSet}{I_{\mathrm{CP}}}
\newcommand{\localIndexSetLabel}[1]{I_{#1}^{\mathrm{local}}}
\newcommand{\representationMap}{\phi}
\newcommand{\kernelFunction}{\kappa}
\newcommand{\localDistance}{d}
\newcommand{\localWeight}{w}
\newcommand{\bandwidth}{h}

\newcommand{\localQuantile}{\widehat{q}_{\mathrm{loc}}}
\newcommand{\localQuantileLabel}[1]{\widehat{q}_{\mathrm{loc},#1}}
\newcommand{\failureRate}{\upalpha}
\newcommand{\correctionQuantile}{\widehat{q}_{\mathrm{corr}}}
\newcommand{\correctionQuantileLabel}[1]{\widehat{q}_{\mathrm{corr},#1}}
\newcommand{\slcpThreshold}{\widehat{q}_{\mathrm{SLCP}}}

\newcommand{\CPpredictionSet}{\widehat{C}}

\title{Localized Conformal Safety Monitoring with Vision-Language Models for Autonomous Driving}
    
\ifarxiv
\author{
\IEEEauthorblockN{Lu\'is Marques, Rong Fang,
Disha Kamale\IEEEauthorrefmark{2}, and
Dmitry Berenson\IEEEauthorrefmark{2}\thanks{
\IEEEauthorrefmark{2}Joint senior authors.}}
\IEEEauthorblockA{University of Michigan, Ann Arbor, MI, USA\\
Email: \texttt{lmarques@umich.edu}}
}
\else
\author{\IEEEauthorblockN{Anonymous Authors}}
\fi

\begin{document}

\bstctlcite{IEEEexample:BSTcontrol}
\maketitle
\renewcommand{\thefootnote}{\fnsymbol{footnote}}
\renewcommand{\thefootnote}{\arabic{footnote}}
\begingroup\color{black}
\renewcommand{\review}[1]{\textcolor{black}{#1}}
\begin{abstract}
Monitoring planned driving trajectories requires accurately estimating the collision likelihood 
with actors whose motion is itself impacted by the ego motion.
Existing classical approaches are often limited by the quality of their forecasting model.
Vision-language models (VLMs) have shown promise in reasoning about the consequences of high-level actions, yet their approximate
predictions are unsuitable for safety-critical applications such as autonomous driving.
Conformal prediction (CP) has emerged as a data-driven framework for quantifying the uncertainty of black-box model predictions.
We propose \methodFullName{},
a post-hoc calibration layer over frozen VLMs that transforms their unreliable predictions into probabilistically calibrated safety prediction sets.
We consider how the ability to estimate safety can depend on the observed driving scene and introduce a localized procedure that upweights 
relevant past experience when calculating uncertainty thresholds. We provide label-conditional finite-sample distribution-free coverage under exchangeability.
Evaluated over 15k CARLA trajectories from unseen scenarios, \methodShortName{} correctly flags
$89.6\%$ of collision-causing trajectories with a Qwen backbone and $88.4\%$ with a
Cosmos backbone, while the base VLMs only flagged $4.6\%$ and $39.1\%$ of the collision-causing trajectories,
respectively. These results indicate that local, label-conditional
calibration can reduce missed unsafe trajectories.
\end{abstract}

\begingroup\color{black}
\endgroup

\section{Introduction and Related Work}
Before self-driving cars execute planned trajectories, they must
assess whether the intended motion can be safely executed, considering the surrounding scene.
This is nontrivial: pedestrians and other vehicles move in response to the ego vehicle.
For example, a trajectory passing a pedestrian near the curb may be
collision-free if the pedestrian waits, but unsafe if they step into the
vehicle's path.
Further, the information available at planning time might miss occluded actors and not provide sufficient signal to enable accurate forecasting of the scene's evolution.
Even seemingly robust driving policies can demonstrate degraded performance when operating in unfamiliar
scenarios --- Fail2Drive reports substantial drops in closed-loop success under
controlled scenario shifts~\cite{gerstenecker2026fail2drive}.

We study the problem of designing a
safety monitor that examines each
candidate trajectory before execution, independently of the quality of the underlying planning module.
That monitor should flag trajectories that would collide within a fixed
time horizon, while avoiding unnecessary alarms hindering task progress.
In general, a safety monitor might need to reason about the candidate
motion and the scene jointly, while considering uncertainty in the true unknown
outcome.
While vision-language driving systems, such as SimLingo, incorporate scene
understanding with action-taking~\cite{renz2025simlingo}, we aim to construct a monitor that can be applied post-hoc over any base planner.
We seek a theoretically-informed way to probabilistically limit the number of missed unsafe trajectories.

\review{\textit{Safety monitors.}\enspace Driving pipelines often forecast the motion of nearby actors
before checking a planned trajectory for collision~\cite{paden2016survey}.
Model-based safety checks can account for uncertain actor motion and
dynamics~\cite{lindemann2023safe}, but their assurances
depend on assumed scene knowledge. When an actor is missed entirely or its response to the
ego vehicle is incorrectly predicted, using a collision check may fail.
We thus investigate the complementary problem of assessing candidate trajectories using
front-facing images, so that a vision-language model (VLM) can assess hard to model risks.
A VLM's raw \safeLabel{}/\unsafeLabel{} prediction provides no
assurances of low-rate missed collisions, since VLM spatial reasoning can be
unreliable~\cite{tian2025drivevlm}.} Hence, we aim to augment VLMs' common-sense reasoning with a post-hoc calibration layer providing probabilistic
guarantees on their prediction performance.

\begin{figure}[!t]
    \centering
    \includegraphics[width=\columnwidth]{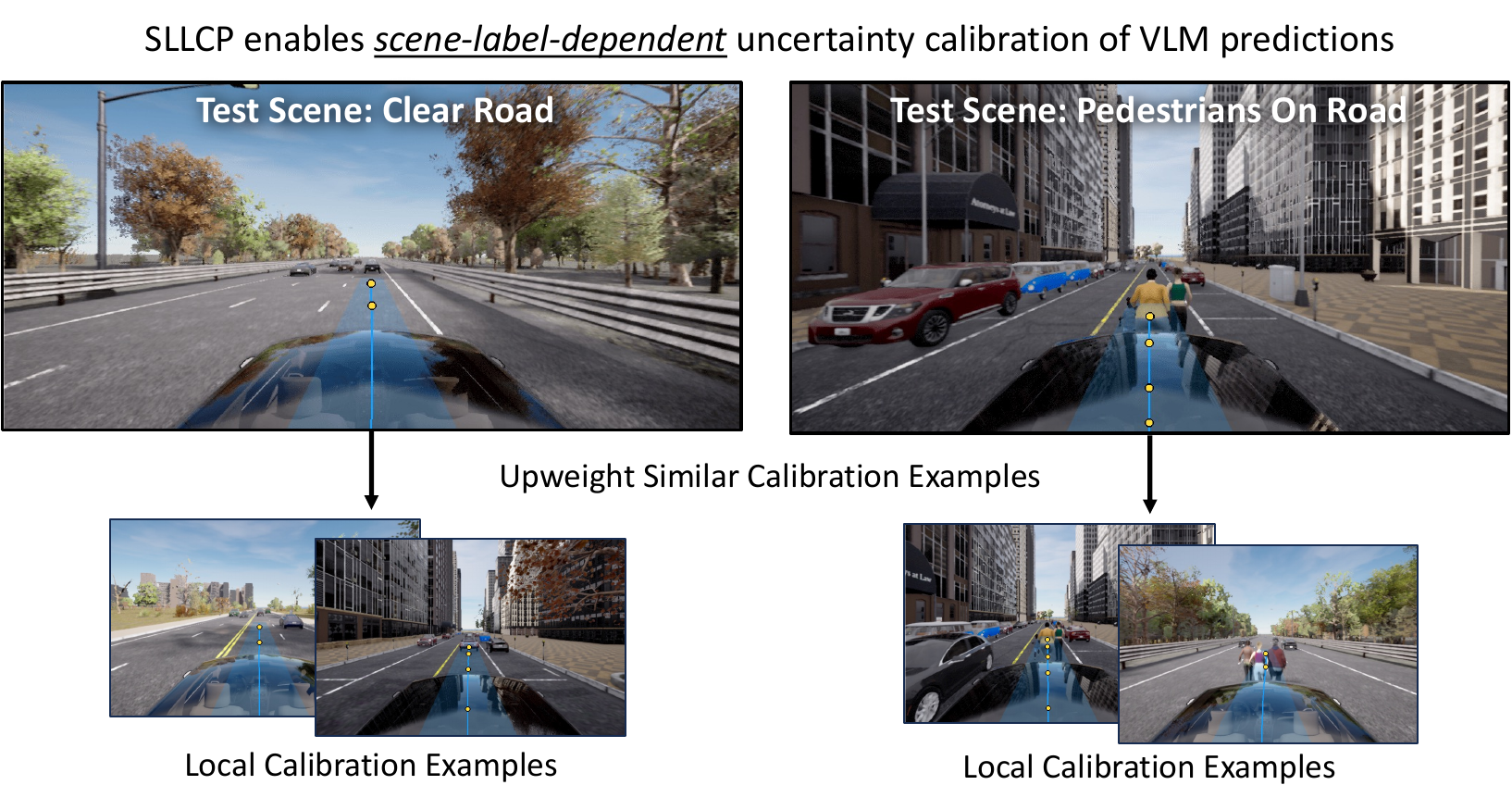}
    \vspace{-22pt}
    \caption{Uncertainty in VLM-based trajectory safety predictions (\safeLabel{} or \unsafeLabel{}) varies across driving scenes. \methodShortName{} upweights calibration examples (bottom) near each test input (top) in the projected embedding space and calibrates each safety label separately to address class imbalance in the collected data (e.g., due to \unsafeLabel{} cases being rarer). Trajectory footprints are shown in blue, with yellow waypoints spaced at $0.5\,$s intervals.}
    \label{fig:title-figure}
    \vspace{-14pt}
\end{figure}

\textit{Conformal prediction in robotics.}\enspace Conformal prediction
(CP) transforms approximate predictions into prediction sets with finite-sample guarantees under
exchangeability~\cite{vovk2005algorithmic,angelopoulos2023gentle}. In
robotics, CP has calibrated: forecasts in dynamic scenes for social navigation
~\cite{lindemann2023safe,dixit2023adaptive}; how dynamics uncertainty varies over a system's
state--action space~\cite{marques2026lucca}; action uncertainty for Lie-algebraic systems~\cite{marques2026claps}; and motion uncertainty across
contact modes~\cite{marques2026capture}.

In this extended abstract, we propose a monitor calibrating a base VLM safety prediction in a scene-adaptive manner.
\review{Our key contributions are:}

\begin{itemize}
    \item We develop \methodFullName{}, a post-hoc method over a planner-agnostic generated trajectory image and a frozen VLM backbone, that
    provides probabilistic guarantees on a low probability of missing unsafe trajectories.
    
    \item We empirically validate the tradeoff between missed unsafe trajectories and
    false alarms on edge-case CARLA scenarios, demonstrating significant improvements in the calibration of two open-weight VLMs, using the same hyperparameter set.
\end{itemize}

\endgroup
\section{Problem Statement}\label{sec:prob-statement}

\looseness-1 Let $\state_t$ denote the state of an autonomous driving agent at time $t\in \mathbb N$, whose planner generates candidate $\horizon$-step-long trajectories $\tau:=\{\state_t, \state_{t+1}, \ldots, \state_{t+\horizon-1}\}$.
We consider the problem of provably monitoring the safety of said finite-horizon trajectories given a black-box planner of \textit{unknown} quality.
We assume that executing $\tau$ without replanning can produce one of two, initially \textit{unknown}, outcomes $\CPoutput$: either a collision occurs within
$\horizon$ steps ($\CPoutput=\unsafeLabel$), or it does not ($\CPoutput=\safeLabel$).
While conventional collision-based monitors could be implemented by projecting $\tau$ into the environment observed so far and performing geometric checks,
the validity of their predictions is dependent on an accurate forecast of the motion of other dynamic entities in the scene,
such as pedestrians and other vehicles.
However, these forecasts are notoriously difficult to estimate accurately, even more so given that the ego motion can influence the behaviours of nearby actors.
In this work, we consider vision-language models (VLMs) with fixed parameters as the base of our safety monitor, due to their common-sense reasoning capabilities.
However, VLM-estimated safety likelihood $\predProb(\CPoutput \mid \CPinput)$ is not directly suitable for safety-critical applications, since it might be overconfident.
Our VLM-based monitor receives as input $\CPinput$, consisting of a front-facing RGB camera image of the scene overlaid with the proposed trajectory,
as well as a prompt describing the safety monitoring task.
\problem{Given a user-specified acceptable failure rate $\failureRate \in (0,1)$, we seek a \textit{set-valued monitor} $\monitor :\CPinputSpace \to2^{\{\safeLabel,\unsafeLabel\}}$
whose probability of missing an unsafe trajectory is at most $\failureRate$.
Formally, we require
\begin{equation}\label{eq:unsafe-miscoverage}
    \mathbb P (\unsafeLabel \notin \CPpredictionSet(\CPinput) \mid \CPoutput = \unsafeLabel) \leq \failureRate,
\end{equation}
and further want to minimize the false positive rate (FPR) 
\begin{equation}\label{eq:fpr}
    \operatorname{FPR} = \mathbb P (\unsafeLabel \in \CPpredictionSet(\CPinput) \mid \CPoutput = \safeLabel),
\end{equation}
since high FPR implies unnecessary alarms which can hinder the agent's task performance.}
To make this problem tractable, we make the standard assumption in the conformal prediction literature of having access to a held-out calibration set
of labeled examples that are iid with the test data, i.e.:
\looseness-1 \begin{assumption}\label{ass:exchangeability}
    We have VLM inputs-outcome pairs $\calibset = \{(\CPinput_i, \CPoutput_i)\}_{i=1}^{\numcal}$ that are iid with
    the test pair $(\CPinputTest, \CPoutputTest)$.
\end{assumption}
Following literature, $\calibset$ is uncorrupted.

\section{Background: Conformal Prediction}\label{sec:cp-background-draft}
We briefly introduce relevant conformal prediction (CP) concepts below and refer to~\cite{vovk2005algorithmic,angelopoulos2023gentle} for deeper insight.
Let $\CPinput\in\CPinputSpace$ denote the available information at the time of prediction and
$\CPoutput\in\CPoutputSpace=\{\safeLabel,\unsafeLabel\}$ the unknown true label.
Given a fixed classifier, the estimated unsafe probability is $\CPpred(x)=\predProb(\unsafeLabel\mid x)$.
The \textit{nonconformity score} function $\scoreFunc:[0,1]\times\CPoutputSpace\to\mathbb R$
measures disagreement between a model prediction and the true label, with higher values indicating higher prediction uncertainty.
For binary classification, a conventional choice is $\scoreFunc(\CPpred(x),\CPcandidate)=1-\predProb(\CPcandidate\mid x)$.

\textit{Split conformal prediction (SplitCP).} SplitCP is a computationally efficient CP variant that uses an exchangeable 
calibration set of labeled examples (Ass.~\ref{ass:exchangeability}) to provide probabilistic guarantees on test-time prediction uncertainty. 
Given $\calibset$, we can evaluate the nonconformity score on each example to obtain a list of scores $\{\scoreVal_i\}_{i=1}^{\numcal}$, where
$\scoreVal_i=\scoreFunc(\CPpred(\CPinput_i),\CPoutput_i)$.
Then the $k=\lceil(\numcal+1)(1-\failureRate)\rceil$th
smallest score in $\{\scoreVal_i\}_{i=1}^{\numcal}\cup\{\infty\}$, serves as a probabilistic threshold $\widehat q$ for the test-time nonconformity score $\scoreFunc(\CPpred(\CPinputTest),\CPoutputTest)$.
Using exchangeability \cite{vovk2005algorithmic}, the following prediction set
\begin{equation}\label{eq:bg-split-set}
\CPpredictionSet(x)=\{\CPcandidate\in\CPoutputSpace:\scoreFunc(\CPpred(x),\CPcandidate)\leq\widehat q\},
\end{equation}
is guaranteed to provide \textit{marginal coverage}, i.e., satisfy $\mathbb P\{\CPoutput\in\CPpredictionSet(\CPinput)\}\geq1-\failureRate$ on average.

\textit{Label-conditional prediction.} Marginal coverage alone does not
account for imbalanced datasets. Since the guarantees are based on average uncertainty, underrepresented labels in $\calibset$ that present higher predictive uncertainty can be
severely undercovered.
For instance, $\calibset$ may contain many more \safeLabel{} than \unsafeLabel{} examples, and $\unsafeLabel$ might be more difficult for the VLM to estimate.
To address this issue, Label-conditional CP instead calibrates each class label separately~\cite{vovk2005algorithmic}.
Let
$I_\CPcandidate=\{i:\CPoutput_i=\CPcandidate\}$ be the calibration indices for class $\CPcandidate$,
$n_\CPcandidate=|I_\CPcandidate|$ the number of calibration examples $\CPcandidate$, and let $\widehat q_\CPcandidate$ be the
$\lceil(n_\CPcandidate+1)(1-\failureRate)\rceil$th smallest score in
$\{\scoreVal_i:i\in I_\CPcandidate\}\cup\{\infty\}$.
Replacing the SplitCP label-shared threshold $\widehat q$ in Eq.~\eqref{eq:bg-split-set} by the per-label threshold $\widehat q_\CPcandidate$ gives rise to the prediction set
\begin{equation}\label{eq:bg-label-set}
\begin{aligned}
\CPpredictionSet(x)&=\{\CPcandidate\in\CPoutputSpace:\scoreFunc(\CPpred(x),\CPcandidate)\leq\widehat q_\CPcandidate\},
\end{aligned}
\end{equation}
which satisfies \textit{label-conditional coverage}, i.e., implies Eq.~\eqref{eq:unsafe-miscoverage}.
This guarantee holds for \textit{finite-sized} $\calibset$ in a \textit{distribution-free} manner, i.e., without making any assumptions on the underlying score distribution.
The prediction region $\CPpredictionSet$ may now contain \safeLabel{}, \unsafeLabel{}, both labels, or the empty set. If $\CPpredictionSet$ includes
\unsafeLabel{}, the monitor should raise a safety alarm.

\textit{Split localized conformal prediction (SLCP).} While each label-dependent threshold $\widehat q_\CPcandidate$ is input-independent,
prediction difficulty may vary across inputs $x$.
For example, some driving scenes may be simpler to classify as unsafe than others, depending on the size, color, and proximity of nearby obstacles.
Hence, we desire a threshold that can adapt to the model uncertainty at each input.
Unfortunately, input-conditional coverage $\mathbb P(Y\in\CPpredictionSet(X)\mid X=x)\geq1-\failureRate$ has been shown to be impossible with finite data,
without making additional distributional assumptions~\cite{lei2014distributionfree}.
SLCP aims to approximate input-conditional coverage by fitting a local score quantile $\localQuantile(x)$ which up-weights calibration
examples with inputs ``near'' the test-time input $x$.
The key idea is that nearby inputs might share similar prediction difficulty, and thus their scores should be considered as more informative than the scores
of faraway calibration examples.
Since on its own $\localQuantile(x)$ does not provide any finite-sample guarantees, SLCP
then adds a held-out correction $\correctionQuantile$
to obtain a final composed threshold $\slcpThreshold(x)$~\cite{guan2023localized,han2022split}.
For exchangeability to hold, $\calibset$ must be initially split into two disjoint sets $\localCalSet$ and $\cpCalSet$ used for local fitting and held-out correction, respectively.
Let $\localIndexSet$ and $\cpIndexSet$ index $\localCalSet$ and $\cpCalSet$, respectively, and let the local distance between a calibration input $\CPinput_i$ and a test input $x$ be
$d_i(x)=\localDistance(\representationMap(\CPinput_i),\representationMap(x))$ where $\representationMap$ is a feature mapping.
A kernel $\kernelFunction$ and bandwidth $\bandwidth>0$ define the local weights
\begin{equation}\label{eq:bg-local-weights}
\localWeight_i(x)=
\frac{\kernelFunction(d_i(x)/\bandwidth)}
{\sum_{j\in \localIndexSet}
\kernelFunction(d_j(x)/\bandwidth)},
\end{equation}
and $\localQuantile(x)$ is the weighted $(1-\failureRate)$ quantile
of $\localCalSet$ scores.
The held-out correction $\correctionQuantile$ is the SplitCP quantile
of $\{\scoreVal_j-\localQuantile(\CPinput_j):j\in\cpIndexSet\}$, giving the SLCP composite threshold of
\begin{equation}\label{eq:bg-slcp-threshold}
\slcpThreshold(x)=
\localQuantile(x)+\correctionQuantile.
\end{equation}
Despite possibly better empirical performance, SLCP only provides \emph{marginal coverage} due to $\correctionQuantile$'s construction~\cite{han2022split}.

\section{Method: \methodShortName{}}\label{sec:method}

\begin{figure}[!t]
    \centering
    \includegraphics[width=\columnwidth]{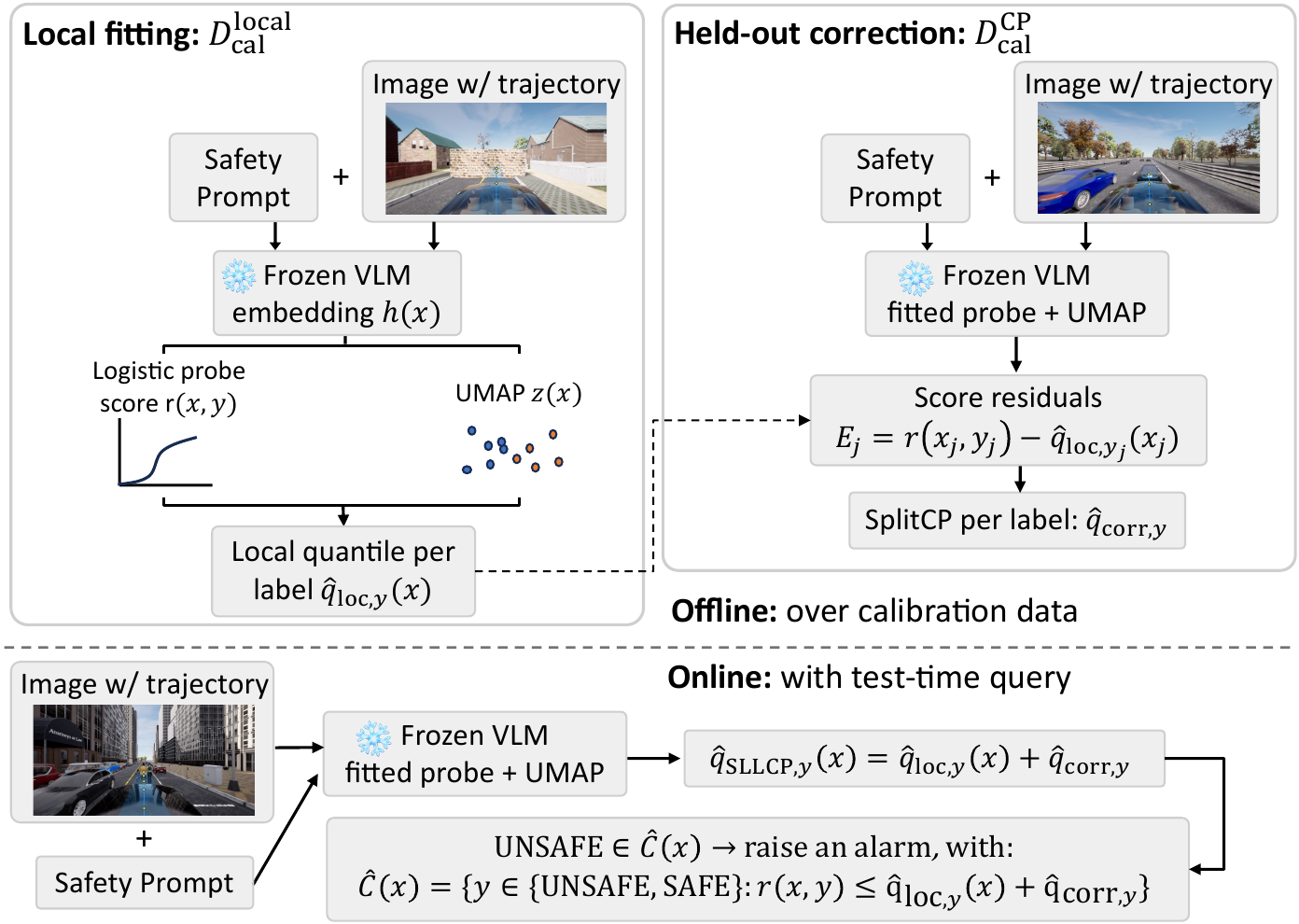}
\looseness-1 \caption{\methodShortName{} overview. Monitor inputs
consist of a camera image with the proposed trajectory overlaid and a fixed safety prompt.
Offline: using $\localCalSet$, we fit a logistic probe to obtain safety scores and a UMAP projection for localization, then define per-label local score quantiles.
The disjoint $\cpCalSet$ provides per-label corrections from score residuals.
Online: given a test-time image--prompt pair, we combine its
local quantiles with the fitted corrections to construct
$\CPpredictionSet(x)$.
The monitor raises an alarm when
$\unsafeLabel\in\CPpredictionSet(x)$.}
\vspace{-4mm}
    \label{fig:method-pipeline}
\end{figure}

We propose \methodFullName{}, which builds on SLCP by combining it with label-conditional conformal prediction. Instead of using
the raw VLM nonconformity score, we learn a nonconformity score to better inform the local quantile fit.
Figure~\ref{fig:method-pipeline} summarizes our method. Following SLCP, we split $\calibset=\localCalSet\sqcup\cpCalSet$, using
$\localCalSet$ to both learn $r$ and fit $\localQuantile$, while $\cpCalSet$ now informs per-label
conformal corrections $\correctionQuantileLabel{\CPcandidate}$ for each label $\CPcandidate\in\CPoutputSpace$.

\runin{Probe score and local representation with $\localCalSet$}
Let $\vlmEmbedding(x)$ be the VLM's hidden state before the Boolean answer token ($\safeLabel$, $\unsafeLabel$) and after the
frozen language model's final RMSNorm.
A raw VLM-based score could be computed from the VLM's Boolean-token
probabilities.
However, we have empirically observed that this score may not expose all safety-relevant information contained in the VLM hidden
state.
Therefore, we fit a linear logistic probe $\CPpred(x)=\sigma(\mathbf w^\top\vlmEmbedding(x)+b)$ to $\vlmEmbedding(x)$ using the
ground-truth labels in $\localCalSet$.
This simple classifier provides an alternative way to estimate the unsafe probability $\CPpred(x)$, which we then use
instead of Boolean token probabilities to compute the standard binary nonconformity score of Sec.~\ref{sec:cp-background-draft}.
This is cheap and does not require updating any VLM weights.
For localization, since the raw VLM embedding is very high-dimensional, we 
first fit a two-dimensional UMAP~\cite{mcinnes2018umap} projection $\projectionMap$ to $\{\vlmEmbedding(\CPinput_i):i\in\localIndexSet\}$, yielding a lower-dimensional latent representation
of the input $\projectedEmbedding(x)=\projectionMap(\vlmEmbedding(x))$.
This $\projectedEmbedding(x)$ is then used as the feature mapping $\representationMap(x)$ in Eq.~\eqref{eq:bg-local-weights}, leading to the local 
distance $\localDistance(u,v)=\|u-v\|_2$ for projected vectors $u:=\projectedEmbedding(\CPinput_i),v:=\projectedEmbedding(x)$.
Both the classifier and UMAP fits are fixed independently of
$\cpCalSet$.

\runin{Label-conditional local quantiles and correction}
For each label $\CPcandidate\in\CPoutputSpace$, let
$\localIndexSetLabel{\CPcandidate}=\{i\in \localIndexSet:\CPoutput_i=\CPcandidate\}$.
We obtain the label-conditional thresholds $\localQuantileLabel{\CPcandidate}(x)$ by computing the weighted
$(1-\failureRate)$ quantile of the scores
$\scoreVal_i=\scoreFunc(\CPpred(\CPinput_i),\CPcandidate)$ for
$i\in\localIndexSetLabel{\CPcandidate}$, using Eq.~\eqref{eq:bg-local-weights}.
Given correction examples $j\in\cpIndexSet$ with
$\CPoutput_j=\CPcandidate$, their residual quantile gives
$\correctionQuantileLabel{\CPcandidate}$ similarly to SLCP. The $\CPpredictionSet$ then is
\begin{equation}\label{eq:method-prediction-set}
\CPpredictionSet(x)=\{\CPcandidate\in\CPoutputSpace:
\scoreFunc(\CPpred(x),\CPcandidate)\leq\localQuantileLabel{\CPcandidate}(x)+\correctionQuantileLabel{\CPcandidate}\}.
\end{equation}
Our safety monitor raises an alarm when $\unsafeLabel\in\CPpredictionSet(x)$, to match the objective in
Sec.~\ref{sec:prob-statement}.
Assuming within-label exchangeability, the proposed method provides label-conditional
marginal coverage.
While this assumption might not hold in our experiments, 
we demonstrate that our theory-driven method still provides adequate empirical coverage in practice.

\section{Experimental Results}\label{sec:exp}
We evaluate if the proposed safety monitor can detect collision-causing trajectories in the CARLA simulator \cite{dosovitskiy2017carla}.
We report unsafe-label coverage, the monitor's false-positive rate on safe trajectories,
and the fraction of produced prediction sets containing both labels. The latter indicates ``ambiguity'' in the monitor's decision,
and could be used to trigger a different recovery policy than $\unsafeLabel$-alarms.
We use $\failureRate=0.1$ for all
experiments.
Coverage is a random variable that, under exchangeability, follows a Beta-binomial distribution given a fixed calibration set and a finite test set~\cite{marquesf2025coverage}.
We report if the observed empirical coverage falls within the $90\%$ confidence interval of the nominal coverage $1-\failureRate$, which would indicate if the applied procedure mitigates the nonexchangeable nature of the data.

We compare \methodShortName{} with several relevant baselines: two uncalibrated raw-VLM rules and four component
ablations. \methodRawBinaryName{} emits one \safeLabel{}/\unsafeLabel{}
decision: the higher-scoring candidate label. \methodRawBandName{} instead uses
a $\failureRate$-aware scoring and a two-label prediction set.
This would be the natural monitor if the raw VLM score were well-calibrated.
The
ablations remove, one at a time: locality
(global, input-independent label-conditional CP); the linear probe (using original VLM score); the UMAP projection
(raw embedding distances); the per-label correction (use label-agnostic uncertainty threshold).
This analysis aims to isolate the impact of our choices of score, latent space representation,
and label-aware correction.

We collect trajectories and front-facing images
using the SimLingo planner~\cite{renz2025simlingo}. A trajectory is deemed
\unsafeLabel{} if tracking it for $2.5\,$s with a PID controller causes a
collision with a vehicle, pedestrian, or other object.
In total, we obtain 40,455 frames from 90 episodes across 30
underlying scenario IDs in three Fail2Drive families~\cite{gerstenecker2026fail2drive}:
\emph{HardBrake}, \emph{PedestriansOnRoad}, and \emph{Wall} (seen in Figure~\ref{fig:method-pipeline}).
As base VLMs, we evaluated
Qwen3-VL-8B-Instruct~\cite{bai2025qwen3vl} and
Cosmos-Reason2-8B~\cite{nvidia2026cosmosreason2}.
We relied on open-weight VLMs so that we could access the hidden embeddings needed for the
linear probe and for doing localized CP in embedding-space.
Neither VLM was fine-tuned and both received the same fixed binary safety prompt and trajectory image.

\looseness-1 We split the entire data into $\calibset$ and an evaluation set.
All seeds of each scenario ID stay on the same side of this split.
This prevents leakage of information from the evaluation set into $\calibset$ ---
each test scenario was not seen during the offline calibration phase.
For some methods, $\calibset$ is further divided into
$\localCalSet$ and $\cpCalSet$. This results in $|\localCalSet|=$ 19,528
frames (2,615 \unsafeLabel{}), $|\cpCalSet|=$ 5,606
frames (856 \unsafeLabel{}), and an evaluation set with 15,321 frames
(2,489 \unsafeLabel{}; 12,832 \safeLabel{}).
\methodShortName~projected VLM embeddings to UMAP-2 (100 neighbors; Euclidean distance), then used a Laplace kernel $\kernelFunction(u)=\exp(-u)$, with per-label bandwidth $\bandwidth_{\CPcandidate} = 0.4$ times the
label-specific median distance in $\localCalSet$.
Hyperparameters were shared across the two VLMs, yet this results in model-specific conformal thresholds.
Table~\ref{tab:direct-umap-results} reports the three evaluation metrics over the 15,321 frames in the evaluation set.

\begin{table}[!t]
\centering\scriptsize\setlength{\tabcolsep}{2pt}\renewcommand{\arraystretch}{1.25}
\caption{Trajectory-safety results on unseen scenario IDs.}
\label{tab:direct-umap-results}
\begin{tabular}{@{}>{\centering\arraybackslash}m{0.28\columnwidth}>{\centering\arraybackslash}m{0.18\columnwidth}>{\centering\arraybackslash}m{0.18\columnwidth}>{\centering\arraybackslash}m{0.13\columnwidth}>{\centering\arraybackslash}m{0.16\columnwidth}@{}}
\hline
\noalign{\vskip 3pt}
Method & VLM & \shortstack{UNSAFE\\coverage (\%) $\uparrow$} & \shortstack{Alarm\\FPR (\%) $\downarrow$} & \shortstack{Ambiguity\\(\%) $\downarrow$} \\
\noalign{\vskip 2pt}
\hline
\multirow{2}{0.28\columnwidth}{\centering \methodRawBinaryName{}} & \includegraphics[height=1.6ex]{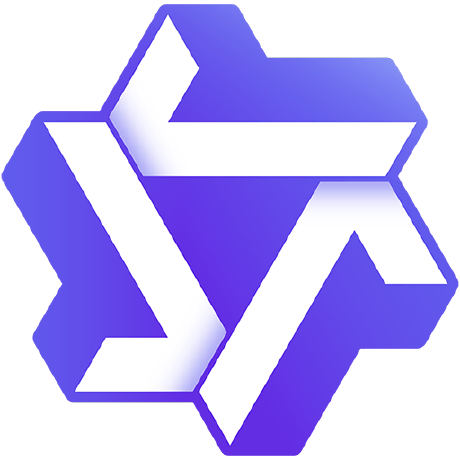}\,Qwen & \cellcolor{red!15}4.6 & 0.3 & 0.0 \\
 & \cellcolor{gray!10}\raisebox{-0.2ex}{\includegraphics[trim=158bp 287bp 230bp 117bp,clip,height=1.7ex]{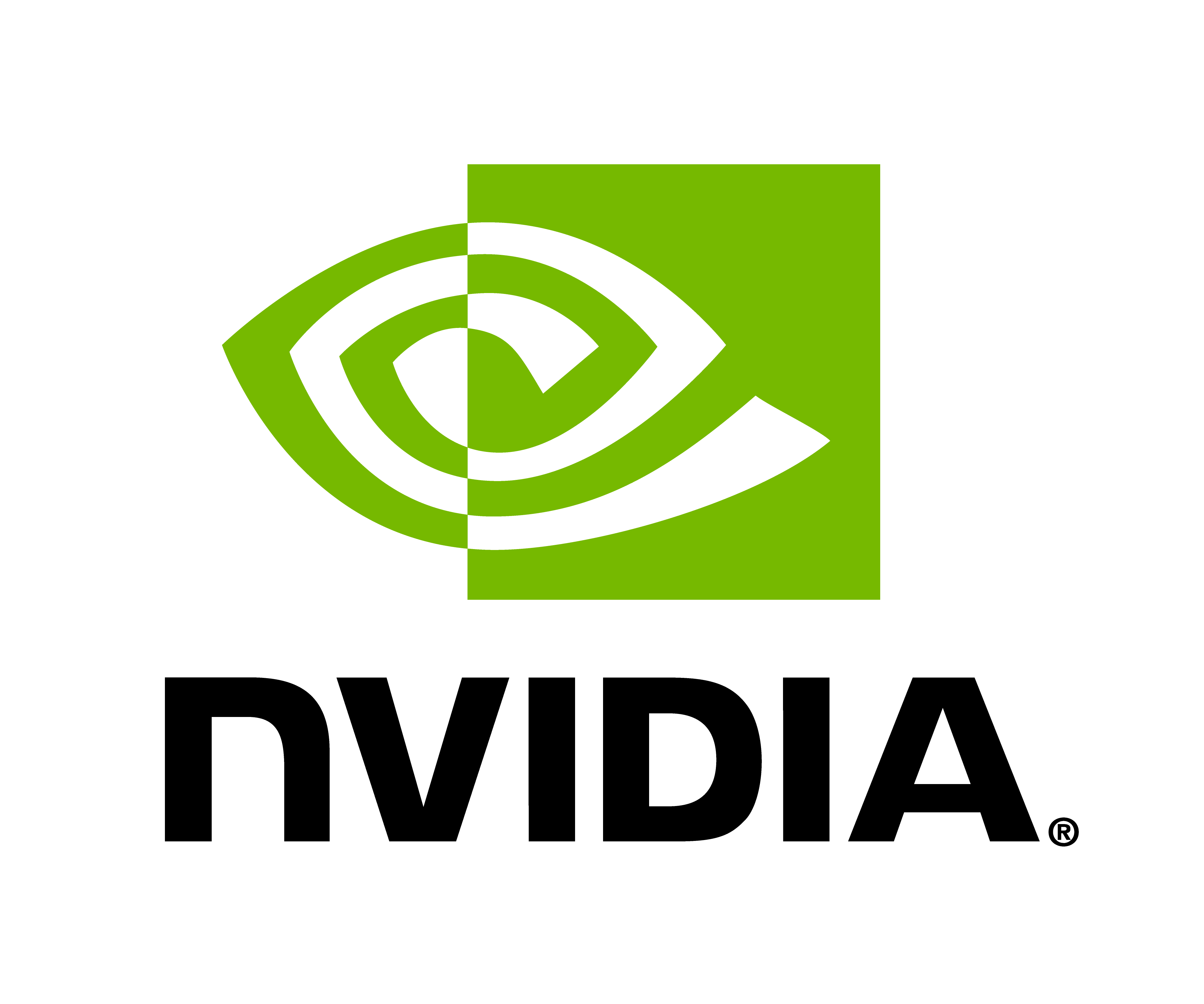}}\,Cosmos & \cellcolor{red!15}39.1 & \cellcolor{gray!10}1.9 & \cellcolor{gray!10}0.0 \\
\hline
\multirow{2}{0.28\columnwidth}{\centering \methodRawBandName{}} & \includegraphics[height=1.6ex]{figures/qwen_logo.png}\,Qwen & \cellcolor{red!15}20.1 & 2.1 & 4.9 \\
 & \cellcolor{gray!10}\raisebox{-0.2ex}{\includegraphics[trim=158bp 287bp 230bp 117bp,clip,height=1.7ex]{figures/nvidia_eye.png}}\,Cosmos & \cellcolor{red!15}78.1 & \cellcolor{gray!10}37.3 & \cellcolor{gray!10}43.9 \\
\hline
\multirow{2}{0.28\columnwidth}{\centering \methodNoLocalityName{}} & \includegraphics[height=1.6ex]{figures/qwen_logo.png}\,Qwen & \cellcolor{red!15}77.5 & 27.7 & 18.6 \\
 & \cellcolor{gray!10}\raisebox{-0.2ex}{\includegraphics[trim=158bp 287bp 230bp 117bp,clip,height=1.7ex]{figures/nvidia_eye.png}}\,Cosmos & \cellcolor{red!15}80.5 & \cellcolor{gray!10}30.4 & \cellcolor{gray!10}21.3 \\
\hline
\multirow{2}{0.28\columnwidth}{\centering \methodNoProbeName{}} & \includegraphics[height=1.6ex]{figures/qwen_logo.png}\,Qwen & \cellcolor{orange!20}99.2 & 95.6 & 89.4 \\
 & \cellcolor{gray!10}\raisebox{-0.2ex}{\includegraphics[trim=158bp 287bp 230bp 117bp,clip,height=1.7ex]{figures/nvidia_eye.png}}\,Cosmos & \cellcolor{orange!20}99.6 & \cellcolor{gray!10}99.0 & \cellcolor{gray!10}89.3 \\
\hline
\multirow{2}{0.28\columnwidth}{\centering \methodNoProjectionName{}} & \includegraphics[height=1.6ex]{figures/qwen_logo.png}\,Qwen & \cellcolor{red!15}84.3 & 41.6 & 28.3 \\
 & \cellcolor{gray!10}\raisebox{-0.2ex}{\includegraphics[trim=158bp 287bp 230bp 117bp,clip,height=1.7ex]{figures/nvidia_eye.png}}\,Cosmos & \cellcolor{red!15}82.4 & \cellcolor{gray!10}37.9 & \cellcolor{gray!10}26.8 \\
\hline
\multirow{2}{0.28\columnwidth}{\centering \methodNoLabelCPName{}} & \includegraphics[height=1.6ex]{figures/qwen_logo.png}\,Qwen & \cellcolor{red!15}70.9 & 15.4 & 6.1 \\
 & \cellcolor{gray!10}\raisebox{-0.2ex}{\includegraphics[trim=158bp 287bp 230bp 117bp,clip,height=1.7ex]{figures/nvidia_eye.png}}\,Cosmos & \cellcolor{red!15}66.7 & \cellcolor{gray!10}11.9 & \cellcolor{gray!10}5.2 \\
\hline
\multirow{2}{0.28\columnwidth}{\centering \textbf{\methodOursName{}}} & \includegraphics[height=1.6ex]{figures/qwen_logo.png}\,Qwen & \cellcolor{green!15}89.6 & 51.8 & 37.6 \\
 & \cellcolor{gray!10}\raisebox{-0.2ex}{\includegraphics[trim=158bp 287bp 230bp 117bp,clip,height=1.7ex]{figures/nvidia_eye.png}}\,Cosmos & \cellcolor{green!15}88.4 & \cellcolor{gray!10}44.5 & \cellcolor{gray!10}33.6 \\
\hline
\end{tabular}
\par\smallskip\noindent
\begin{minipage}{\columnwidth}\scriptsize
All methods use the same 15,321 whole-ID test frames (2,489 UNSAFE; 12,832 SAFE). Ablations remove one component of \methodShortName{}: locality (global label CP), logistic probe (original VLM score), projection (raw VLM embedding), or label-conditional final CP (one threshold shared across labels). The first three retain per-label correction; the last does not. Alarm FPR indicates inclusion of \unsafeLabel{} in $\CPpredictionSet$ on \safeLabel{} frames. Ambiguity is the fraction of frames with $|\CPpredictionSet|=2$. \colorbox{red!15}{red}: coverage below the 90\% Beta--binomial confidence interval (too optimistic); \colorbox{green!15}{green}: within the interval; \colorbox{orange!20}{orange}: above the interval (too conservative).
\vspace{-4mm}
\end{minipage}
\end{table}

\methodRawBinaryName{} alarms included \unsafeLabel{} on only
$4.6\%$ of Qwen's and $39.1\%$ of Cosmos's unsafe frames, failing to meet the user-set safety requirements.
Similarly, the safety-biased
raw rule also remained well below the nominal $90\%$ target. Our approach,
\methodShortName{}, increased unsafe-label coverage to $89.6\%$ and $88.4\%$ staying within the finite-sample Beta-binomial confidence interval of the user-set requirement.
However, the alarm FPRs rose to $51.8\%$ and $44.5\%$ and both-label ambiguity to $37.6\%$
and $33.6\%$, respectively. This indicates an efficiency--coverage tradeoff.
In safety-critical applications, where consequences can be severe, one may prefer to reduce missed unsafe trajectories at the cost of more false alarms.

The ablations motivate our design decisions. Without the logistic probe, the
local method includes nearly all unsafe frames but raises false alarms on at least
$95.6\%$ of safe frames, making it impractical for real-world applications.
Without UMAP, unsafe coverage falls to $84.3\%$ and
$82.4\%$. Using a label-independent threshold lowers coverage further to $70.9\%$ and
$66.7\%$. These comparisons suggest that the score, local
representation, and label-specific correction each impact the
coverage--false-alarm tradeoff.

\looseness-1 \runin{Limitations} Results are sensitive to the UMAP projection and kernel
hyperparameters. The original dataset includes video frames collected along full-episode SimLingo trajectories.
Hence, frames are correlated, breaking exchangeability.
The scenario shift from $\calibset$ also impacts empirical performance.

\vspace{-4mm}
\section{Conclusion}
On unseen Fail2Drive scenarios in CARLA, \methodShortName{} achieved $89.6\%$
and $88.4\%$ unsafe-label coverage for Qwen and Cosmos, improvements of $85.0$ and $49.3$ percentage points, respectively, over the raw VLM performance.
In this extended abstract, we considered trajectory monitoring as a two-label prediction-set problem.
We combined a learned score, from frozen VLM representations, with local calibration in a projected
embedding space and label-conditional calibration.
Compared with the no-probe ablation, our approach reduces the alarm FPR by
$43.8$ and $54.5$ percentage points for Qwen and Cosmos,
respectively, while keeping unsafe-label coverage near $90\%$.
\methodShortName{} provides a planner-agnostic way to balance safety monitoring and 
false alarms without requiring any VLM fine-tuning.
Future work could consider applications to semantic safety violations, such as
failing to yield or obey traffic signals. Yet, obtaining reliable
ground-truth labels in these settings is nontrivial.

\bibliographystyle{IEEEtranN}
\bibliography{references}

@IEEEtranBSTCTL{IEEEexample:BSTcontrol,
  CTLuse_forced_etal       = "yes",
  CTLmax_names_forced_etal = "6",
  CTLnames_show_etal       = "1"
}

@book{vovk2005algorithmic,
  author    = {Vovk, Vladimir and Gammerman, Alex and Shafer, Glenn},
  title     = {Algorithmic Learning in a Random World},
  publisher = {Springer},
  address   = {New York, NY, USA},
  year      = {2005},
  doi       = {10.1007/b106715}
}

@article{angelopoulos2023gentle,
  author  = {Angelopoulos, Anastasios N. and Bates, Stephen},
  title   = {Conformal Prediction: A Gentle Introduction},
  journal = {Found. Trends Mach. Learn.},
  year    = {2023},
  doi     = {10.1561/2200000101}
}

@article{marquesf2025coverage,
  author  = {Paulo C. {Marques F.}},
  title   = {Universal Distribution of the Empirical Coverage in Split Conformal Prediction},
  journal = {Statist. Probab. Lett.},
  year    = {2025},
  doi     = {10.1016/j.spl.2024.110350}
}

@article{lei2014distributionfree,
  author  = {Lei, Jing and Wasserman, Larry},
  title   = {Distribution-Free Prediction Bands for Non-Parametric Regression},
  journal = {J. R. Stat. Soc. Ser. B},
  year    = {2014},
  doi     = {10.1111/rssb.12021}
}

@article{guan2023localized,
  author  = {Guan, Leying},
  title   = {Localized Conformal Prediction:
             A Generalized Inference Framework for Conformal Prediction},
  journal = {Biometrika},
  year    = {2023},
  doi     = {10.1093/biomet/asac040}
}

@misc{han2022split,
  author        = {Han, Xing and Tang, Ziyang and Ghosh, Joydeep and Liu, Qiang},
  title         = {Split Localized Conformal Prediction},
  year          = {2022},
  note          = {arXiv preprint arXiv:2206.13092}
}

@article{lindemann2023safe,
  author  = {Lindemann, Lars and Cleaveland, Matthew and Shim, Gihyun
             and Pappas, George J.},
  title   = {Safe Planning in Dynamic Environments Using Conformal Prediction},
  journal = {IEEE Robot. Autom. Lett.},
  year    = {2023},
  doi     = {10.1109/LRA.2023.3292071}
}

@inproceedings{dixit2023adaptive,
  author    = {Dixit, Anushri and Lindemann, Lars and Wei, Skylar X.
               and Cleaveland, Matthew and Pappas, George J.
               and Burdick, Joel W.},
  title     = {Adaptive Conformal Prediction for Motion Planning
               among Dynamic Agents},
  booktitle = {L4DC},
  year      = {2023}
}

@incollection{marques2026lucca,
  author    = {Marques, Lu{\'i}s and Berenson, Dmitry},
  title     = {Quantifying Aleatoric and Epistemic Dynamics Uncertainty
               via Local Conformal Calibration},
  booktitle = {WAFR},
  year      = {2026},
  doi       = {10.1007/978-3-032-09970-9_5}
}

@article{marques2026claps,
  author  = {Marques, Lu{\'i}s and Ghaffari, Maani and Berenson, Dmitry},
  title   = {Lies We Can Trust: Quantifying Action Uncertainty With Inaccurate Stochastic Dynamics Through Conformalized Nonholonomic {Lie} Groups},
  journal = {IEEE Robot. Autom. Lett.},
  year    = {2026},
  doi     = {10.1109/LRA.2026.3656773}
}

@inproceedings{marques2026capture,
  author    = {Marques, Lu{\'i}s and Popov, Kristian and Berenson, Dmitry},
  title     = {Particle-Based Conformal Prediction for Contact-Aware Uncertainty Calibration in Stratified Configuration Spaces},
  booktitle = {COPA},
  year      = {2026}
}

@article{paden2016survey,
  author  = {Paden, Brian and {\v{C}}{\'a}p, Michal and Yong, Sze Zheng
             and Yershov, Dmitry and Frazzoli, Emilio},
  title   = {A Survey of Motion Planning and Control Techniques
             for Self-Driving Urban Vehicles},
  journal = {IEEE Trans. Intell. Veh.},
  year    = {2016},
  doi     = {10.1109/TIV.2016.2578706}
}

@inproceedings{tian2025drivevlm,
  author    = {Tian, Xiaoyu and Gu, Junru and Li, Bailin and Liu, Yicheng
               and Wang, Yang and Zhao, Zhiyong and Zhan, Kun
               and Jia, Peng and Lang, Xianpeng and Zhao, Hang},
  title     = {{DriveVLM}: The Convergence of Autonomous Driving
               and Large Vision-Language Models},
  booktitle = {CoRL},
  year      = {2025}
}

@inproceedings{dosovitskiy2017carla,
  author    = {Dosovitskiy, Alexey and Ros, German and Codevilla, Felipe
               and Lopez, Antonio and Koltun, Vladlen},
  title     = {{CARLA}: An Open Urban Driving Simulator},
  booktitle = {CoRL},
  year      = {2017}
}

@inproceedings{renz2025simlingo,
  author    = {Renz, Katrin and Chen, Long and Arani, Elahe
               and Sinavski, Oleg},
  title     = {{SimLingo}: Vision-Only Closed-Loop Autonomous Driving
               with Language-Action Alignment},
  booktitle = {CVPR},
  year      = {2025},
  doi       = {10.1109/CVPR52734.2025.01120}
}

@inproceedings{gerstenecker2026fail2drive,
  author    = {Gerstenecker, Simon and Geiger, Andreas and Renz, Katrin},
  title     = {{Fail2Drive}: Benchmarking Closed-Loop Driving Generalization},
  booktitle = {IEEE/RSJ International Conference on Intelligent Robots and Systems (IROS)},
  year      = {2026}
}

@misc{bai2025qwen3vl,
  author        = {Bai, Shuai and Cai, Yuxuan and Chen, Ruizhe and others},
  title         = {{Qwen3-VL} Technical Report},
  year          = {2025},
  note          = {arXiv preprint arXiv:2511.21631}
}

@misc{nvidia2026cosmosreason2,
  author       = {{NVIDIA}},
  title        = {{Cosmos-Reason2}},
  howpublished = {NVIDIA Cosmos documentation},
  year         = {2026},
  note         = {Accessed September 24, 2026}
}

@article{mcinnes2018umap,
  author  = {McInnes, Leland and Healy, John and Saul, Nathaniel
             and Gro{\ss}berger, Lukas},
  title   = {{UMAP}: Uniform Manifold Approximation and Projection},
  journal = {JOSS},
  year    = {2018},
  doi     = {10.21105/joss.00861}
}

\end{document}